\documentclass[sigconf, screen]{acmart}

\AtBeginDocument{%
  }

\setcopyright{acmlicensed}
\copyrightyear{2025}
\acmYear{2025}
\acmDOI{XXXXXXX.XXXXXXX}

\acmConference[ACM MM '25]{Proceedings of the 33nd ACM International Conference
on Multimedia}{October 27--31,
  2025}{Dublin, ireland}

\acmISBN{979-8-4007-XXXX-X/25/10}

\usepackage{graphicx}
\usepackage[T1]{fontenc}

\usepackage{xcolor}
\usepackage{colortbl}
\usepackage{booktabs}
\usepackage{adjustbox}
\usepackage{arydshln}

\usepackage[edges]{forest}
\usepackage{arydshln}
\usepackage{pgfplots}
\usepackage{makecell}

\usepackage{dsfont}
\usepackage{courier}
\usepackage[most]{tcolorbox}
\usepackage{multirow}
\definecolor{gold}{RGB}{234, 51, 35}

\definecolor{titlered}{RGB}{184,49,55}
\newcommand{\titlered}[1]{\textcolor{titlered}{#1}}

\definecolor{titleblue}{RGB}{59,101,155}
\newcommand{\titleblue}[1]{\textcolor{titleblue}{#1}}

\usepackage{arydshln}
\usepackage[most]{tcolorbox}
\usepackage{multirow}
\newtcolorbox{taskbox}[2][]{
	enhanced, breakable,
	colframe=blue3!40,
	colback=blue5!5,
	arc=1mm,
	outer arc=1mm,
	fontupper=\small,
	fontlower=\small,
	coltitle=blue1,
	fonttitle=\bfseries,
	boxsep=1mm,
	left=0mm,
	right=0mm,
	top=0mm,
	bottom=0mm,
	before={\noindent},
	segmentation style={solid, blue3},
	title=#2,
	#1
}

\usepackage{amsmath}
\usepackage{mathtools}
\usepackage{amsfonts}
\usepackage{amssymb}
\usepackage{pifont}
\usepackage{tabularx}
\usepackage{arydshln}
\usepackage{adjustbox}
\usepackage{paralist} 
\usepackage{CJKutf8} 
\usepackage{booktabs} 
\usepackage{subfig}
\usepackage{xcolor}
\usepackage{colortbl}
\usepackage{booktabs}
\usepackage{adjustbox}
\usepackage{arydshln}

\usepackage{tikz}
\usepackage[edges]{forest}

\usepackage{courier}
\usepackage{pgfplots}
\usepackage{makecell}
\usepackage[most]{tcolorbox}
\definecolor{darkolivegreen}{rgb}{0.33, 0.42, 0.18}
\definecolor{my_green}{RGB}{40,154,121}
\definecolor{my_yellow}{RGB}{255,165,0}
\definecolor{my_red}{RGB}{176,46,46}

\usepackage{hyperref}

\definecolor{red}{RGB}{184,49,55}

\definecolor{blue}{RGB}{55,83,156}

\definecolor{green}{RGB}{100,141,63}

\definecolor{vanblue}{RGB}{247,249,255}
\definecolor{vitgreen}{RGB}{245,251,247}

\newtcolorbox{datasetbox}[2][]{
	width=\columnwidth,
	colback = vitgreen, 
	colframe = vitgreen, 
	boxsep=0pt,left=0pt,right=8pt,top=8pt,bottom=8pt,
	fontupper=\linespread{0.9} ,
	title=#2,#1}

\newtcolorbox{vitcot1}[2][]{
	width=\columnwidth,
	colback = vanblue, 
	colframe = vanblue, 
	boxsep=0pt,left=8pt,right=8pt,top=2pt,bottom=4pt,
	fontupper=\linespread{0.9} ,
	title=#2,#1}

\newtcolorbox{vitcot2}[2][]{
	width=\columnwidth,
	colback = vitgreen, 
	colframe = vitgreen, 
	boxsep=0pt,left=8pt,right=8pt,top=2pt,bottom=0pt,
	fontupper=\linespread{0.9} ,
	title=#2,#1}

\newtcolorbox{vanbox}[2][]{
    width=\columnwidth,
    colback = vanblue,        
    colframe = vanblue,       
    boxsep=0pt,
    left=-8pt, right=8pt,
    top=8pt, bottom=8pt,
    fontupper=\linespread{0.9},
    title=#2,
    #1
}

\newcommand\Video[1]{\colorbox[RGB]{224,235,247}{#1}}

\newcommand\Question[1]{\colorbox[RGB]{243,243,243}{#1}}
\newcommand\InitialReasoning[1]{\colorbox[RGB]{227,238,225}{#1}}

\newtcolorbox{vitbox}[2][]{
    width=\columnwidth,
    colback = vitgreen,
    colframe = vitgreen,
    boxsep=0pt,
    left=-8pt, right=8pt,
    top=8pt, bottom=8pt,
    fontupper=\linespread{0.9},
    title=#2,
    #1
}

\definecolor{myblue}{RGB}{215,226,240}
\definecolor{mygreen}{RGB}{229,238,226}

\begin{document}

\title[\titleblue{\textbf{V}}i\titlered{\textbf{T}}CoT:  \titleblue{\textbf{Video}}-\titlered{\textbf{Text}} Interleaved Chain-of-Thought for \\ Boosting Video Understanding in Large Language Models] {\titleblue{\textbf{V}}i\titlered{\textbf{T}}CoT: {\raisebox{-0.3ex}{\includegraphics[height={1.2\ht\strutbox}]{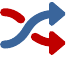}}} \titleblue{\textbf{Video}}-\titlered{\textbf{Text}} Interleaved Chain-of-Thought for Boosting Video Understanding in Large Language Models}

\author{Yongheng Zhang}
\authornote{Both authors contributed equally to this work.}
\author{Xu Liu}
\email{{zyhbrz, xuliu6160}@gmail.com}
\authornotemark[1]
\affiliation{%
  \institution{School of Computer Science and Engineering, Central South University}
  \city{ChangSha}
  \state{Hunan}
  \country{China}
}

\author{Ruihan Tao}
\email{aarontao757@gmail.com}
\affiliation{%
  \institution{School of Computer Science and Engineering, Central South University}
  \city{ChangSha}
  \state{Hunan}
  \country{China}
}

\author{Qiguang Chen}
\email{qgchen@ir.hit.edu.cn}
\affiliation{%
  \institution{Research Center for SCIR,\\ Harbin Institute of Technology}
  \city{Harbin}
  \state{Heilongjiang}
  \country{China}}

\author{Hao Fei}
\email{haofei37@nus.edu.sg}
\affiliation{%
 \institution{NExT Research Center,\\ National
University of Singapore}
 \country{Singapore}}

\author{Wanxiang Che}
\email{car@ir.hit.edu.cn}
\affiliation{%
  \institution{Research Center for SCIR,\\ Harbin Institute of Technology}
  \city{Harbin}
  \state{Heilongjiang}
  \country{China}}

\author{Libo Qin}
\email{lbqin@csu.edu.cn}
\authornote{Corresponding Author.}
\affiliation{%
  \institution{School of Computer Science and Engineering, Central South University}
  \city{ChangSha}
  \state{Hunan}
  \country{China}
}

\renewcommand{\shortauthors}{Yongheng Zhang, Xu Liu, Ruihan Tao, Qiguang Chen, Hao Fei, Wanxiang Che, Libo Qin}

\begin{abstract}
Video understanding plays a vital role in bridging low-level visual signals with high-level cognitive reasoning, and is fundamental to applications such as autonomous driving, embodied AI, and the broader pursuit of AGI. The rapid development of large language models (LLMs), particularly those utilizing Chain-of-Thought (CoT) technology, has significantly advanced video reasoning capabilities. However, current approaches primarily depend on textual information for reasoning, overlooking the visual modality in the actual video reasoning process. In contrast, humans naturally re-examine visual content while reasoning. Motivated by this, we introduce a novel video reasoning paradigm: Video-Text Interleaved CoT (ViTCoT), which facilitates more intuitive and cognitively aligned reasoning. To the end, first, we construct the Video-Text Interleaved Benchmark (ViTIB), which is created using MLLMs for key-video selection and manually verified. Furthermore, we extensively explore the potential of the ViTCoT paradigm in the video understanding field. Extensive experiments demonstrate that ViTCoT significantly boosts performance compared to the traditional text-only CoT paradigm and effectively activates more neuron values in MLLMs.

\end{abstract}



\begin{teaserfigure}
  \includegraphics[width=\textwidth]{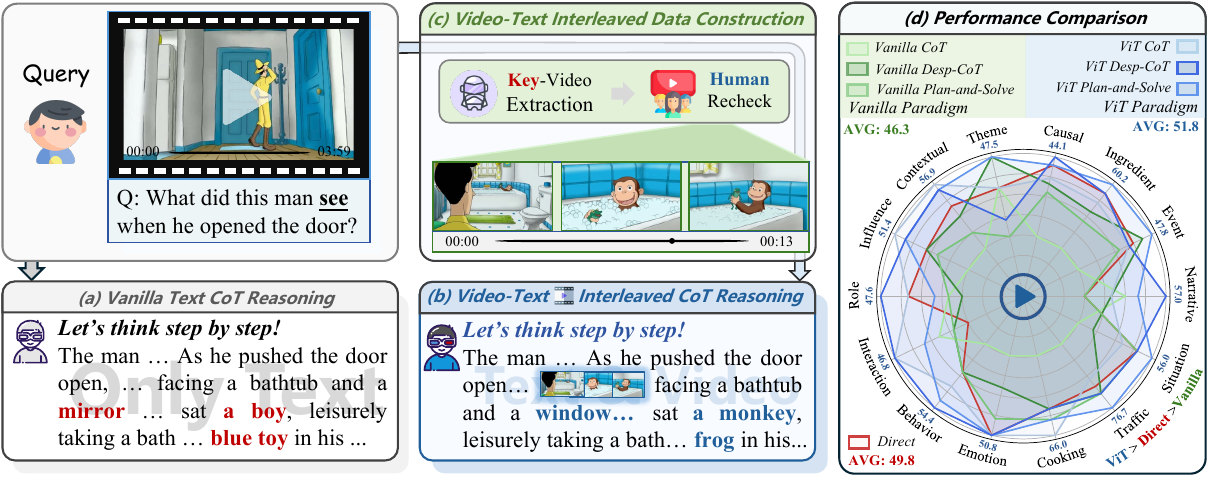}
  \vspace{-8mm}
  \caption{(a) Vanilla Text CoT Reasoning; (b) Video-Text Interleaved CoT Reasoning; (c) Video-Text Interleaved Data Construction; (d) Performance Comparison: Vanilla Reasoning Paradigm (\textit{Vanilla CoT} \cite{wei2022chain}, \textit{Vanilla Desp-CoT} \cite{wu2023role}, and \textit{Vanilla Plan-and-Solve}~\cite{wang2023plan}) vs. Video-Text Interleaved Reasoning Paradigm (\textit{ViT CoT}, \textit{ViT Desp-CoT} and \textit{ViT Plan-and-Solve}) on \textit{Qwen2.5-VL-7B}.}
  \label{fig:intro}
\end{teaserfigure}

\maketitle

\section{Introduction}

Video understanding tasks refer to the automated interpretation and extraction of semantic and contextual information from video data through the analysis of visual, auditory, and temporal modalities~~\cite{tang2023video,cheng2024videgothink,NEURIPS2024_5f280960,xing2024survey,wu2021towards}. These tasks leverage multiple modalities to link low-level signals with high-level cognition, laying the groundwork for understanding real-world phenomena~\cite{lavee2009understanding,NEURIPS2024_5f280960,zhou2024mlvu,wang2024s3,zhang2025cchall}. Therefore, video understanding is widely applied in autonomous driving and embodied Artificial Intelligence, serving as an essential pathway toward achieving Artificial General Intelligence (AGI)~\cite{ying2024mmt, Cui_2024_WACV,cho2024sora,wang2025artragretrievalaugmentedgenerationstructured}, which attracts growing research interest~\cite{maaz-etal-2024-video,zhou2024survey, chen2024sharegpt4video,wu2024pastnet,Wang_2023_ICCV,cheng2025visual}. 

\vspace{0.5mm}

In recent years, with advancements in technology, multimodal large language models (MLLMs) have rapidly developed, significantly advancing video understanding tasks~\cite{weng2024longvlm,team2024gemini,achiam2023gpt,qin2025survey,chen2025ai4research}. Additionally, Chain-of-Thought (CoT) reasoning techniques~\cite{wei2022chain,zhang2024autocap,zhang2024autocap,chen2025towards,cheng2025comt,chen2024unlocking,qinfactors,xu2024activerag}, which encourage models to reason step-by-step, have shown considerable potential and have been validated as essential tools for enhancing LLMs' performance in complex reasoning tasks~\cite{yao2023tree, xia2024beyond, wang2022self}. Based on this, researchers increasingly explore the application of these techniques in video understanding tasks~\cite{fei2024video, wang2024videocot, wang2025multimodal}.
 Specifically, Fei et al. \cite{fei2024video} propose Video-of-Thought, which extends CoT by decomposing complex tasks into simpler steps, from pixel-level perception to high-level reasoning. Wang et al. \cite{wang2024videocot} introduce the VideoCoT benchmark to explore the reasoning limits of MLLMs on multimodal tasks. Yue et al. \cite{yue2025find} leverage dense video text retrieval to extract fine-grained information from long videos, enhancing MLLM prompting performance.

While prior works have achieved impressive results, we find that they rely solely on textual information for reasoning. As shown in Figure~\ref{fig:intro} (a), existing CoT typically processes both video and text inputs but performs reasoning solely based on text.
However, as the adage ``\textbf{A picture is worth a thousand words}'' suggests, visual information holds significant cognitive value in enhancing reasoning. As shown in Figure~\ref{fig:intro}~(b), humans naturally enhance their reasoning during video comprehension by reviewing key visual information, which facilitates a more intuitive cognitive process. 
Unfortunately, this intuitive and cognitively aligned reasoning process has been consistently overlooked in existing research.

Motivated by this, we introduce a novel video reasoning paradigm: \titleblue{\textbf{Video}}-\titlered{\textbf{Text}} Interleaved CoT (\titleblue{\textbf{V}}i\titlered{\textbf{T}}CoT), and aim to achieve the human process of reviewing visual information to realize natural, intuitive reasoning.
Specifically, we research the following: 
(1)~\textbf{\titleblue{Video}-\titlered{Text} Interleaved Benchmark (\titleblue{\textbf{V}}i\titlered{\textbf{T}}IB) Construction}: As shown in Figure~\ref{fig:intro}~(c), to fill the current gap in video understanding research, we develop a Video-Text Interleaved Benchmark. First, we use MLLMs to filter out the key video frames that can assist reasoning, followed by rigorous manual verification through repeated checks. This benchmark serves as a crucial and essential data resource for evaluating MLLMs that integrate video-text reasoning. (2) \textbf{\titleblue{Video}-\titlered{Text} Interleaved CoT (\titleblue{\textbf{V}}i\titlered{\textbf{T}}CoT) Paradigm Exploring}: As shown in Figure~\ref{fig:intro} (b), we explore the Video-Text Interleaved Chain-of-Thought paradigm. By inputting Oracle key-video and interleaving them within the text video reasoning, we enable MLLMs to simulate a human-like analysis of key-video. Through this exploration, we aim to bridge the gap between the current video understanding paradigm and more intuitive, human-like reasoning.

Extensive experiments demonstrate that ViTCoT significantly outperforms traditional text-only CoT in reasoning performance on complex video tasks. As shown in Figure~\ref{fig:intro} (d), ViTCoT, when implemented with \textit{Qwen2.5-VL-7B}, achieves an average improvement of 5.5\% compared to baseline CoT methods. Furthermore, deeper experiments show that the video-text interleaved reasoning paradigm effectively activates more neuron values in MLLMs.

The key contributions of this work are:
\begin{itemize}
	\item We point out that the existing video reasoning paradigm only narrowly relies on the textual modality, which motivates us to rethink the current progress of video understanding reasoning and introduce a novel paradigm: \titleblue{\textbf{Video}}-\titlered{\textbf{Text}} Interleaved CoT (\titleblue{\textbf{V}}i\titlered{\textbf{T}}CoT) video reasoning paradigm.
	\item To fill the research gap, we first develop the \titleblue{\textbf{Video}}-\titlered{\textbf{Text}} Interleaved Benchmark (\titleblue{\textbf{V}}i\titlered{\textbf{T}}IB). Furthermore, we conduct extensive explorations of the \titleblue{\textbf{V}}i\titlered{\textbf{T}}CoT on \titleblue{\textbf{V}}i\titlered{\textbf{T}}IB.
	\item Through experimental validation, it is demonstrated that the video-text interleaved reasoning paradigm can significantly boost the performance of MLLMs in video understanding, effectively activating more neuron values.
\end{itemize}

To facilitate further research, data and code will be open-sourced and publicly available at \url{https://github.com/BRZ911/ViTCoT}.

\section{Background}

\begin{figure*}[t]
	\centering
  \includegraphics[width=\textwidth]{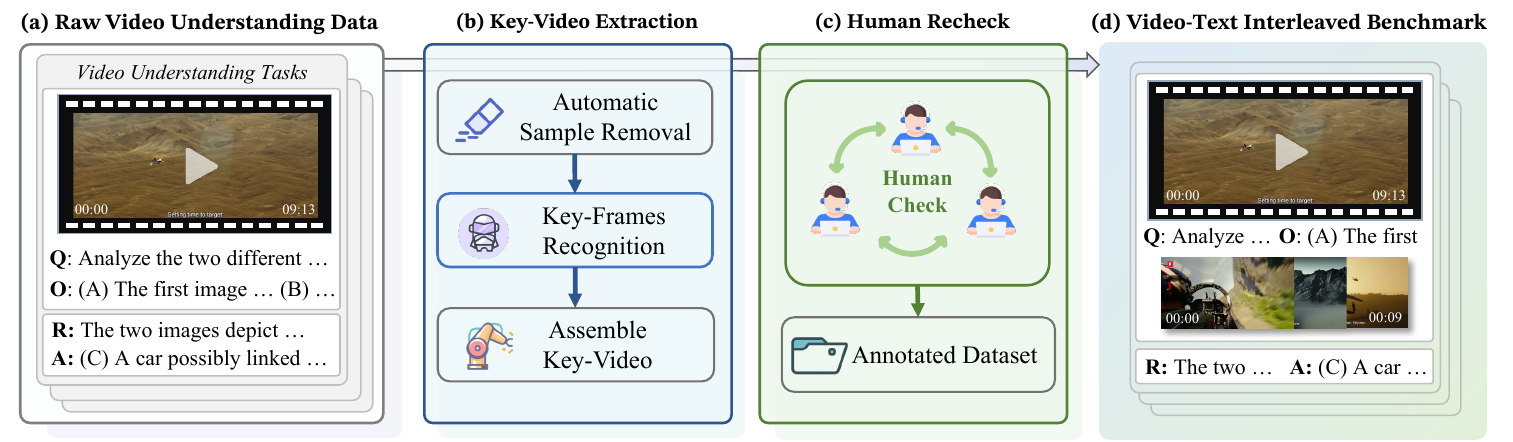}
  \caption{The construction of the Video-Text Interleaved Benchmark includes: (a) \textit{Raw Video Understanding Data}, (b) \textit{Key-Video Extraction}, (c) \textit{Human Recheck}, (d) \textit{Video-Text Interleaved Benchmark}. \textbf{Q}: Question; \textbf{O}: Option; \textbf{R}: Reasoning Process; \textbf{A}: Answer.}
  \label{fig:dataset}
\end{figure*}

\subsection{Vanilla CoT Reasoning Paradigm for \\ Video Understanding}

The Vanilla Chain-of-Thought for Video Understanding \cite{fei2024video} refers to the process in MLLMs where video data, related questions, and reasoning instructions are input into the model for video understanding and reasoning tasks. Specifically, as shown in Figure~\ref{fig:intro}~(a), the MLLMs generate a reasoning process through a series of text-based reasoning steps based on the input video $V$, questions $Q$, and prompt $P_t$, which is denoted as: 

\begin{equation}
	\mathcal{R}_t = \underset{\mathcal{R'}}{\operatorname{argmax}} \ P(\mathcal{R'}\ |V, Q, P_t),
\end{equation}
where $\mathcal{R}_t$ represents the detailed text reasoning process of vanilla reasoning, and $P(\mathcal{R'}\ |V, Q, P_t)$ denotes the probability of accurately generating the reasoning process $\mathcal{R}$ given the input $V$, $Q$ and $P_t$.

\subsection{Video-Text Interleaved CoT Paradigm for Video Understanding}
In contrast to Vanilla CoT, Video-Text Interleaved CoT directly incorporates the video into the reasoning process for auxiliary reasoning. Specifically, as shown in Figure~\ref{fig:intro}~(b), Video-Text Interleaved CoT is divided into two stages. The first stage involves performing preliminary text reasoning, with inputs including video $V$, questions $Q$, and prompt $P_t$, which is denoted as:
\begin{equation}
	\mathcal{R'}_t = \underset{\mathcal{R'}}{\operatorname{argmax}} \ P(\mathcal{R'}\ |V, Q, P_t),
\end{equation}
where $\mathcal{R'}_t$ represents the preliminary reasoning obtained, which is then combined with the key video $V_k \in V$, and video-text interleaved prompt $P_{t \rightarrow v}$ as input to produce the final reasoning:
\begin{equation}
	\mathcal{R}_{t \rightarrow v} = \underset{\mathcal{R}}{\operatorname{argmax}} \ P(\mathcal{R}\ |V_k \oplus \mathcal{R'}_t, V, Q, P_{t \rightarrow v}),
\end{equation}
where $\mathcal{R}_{t \rightarrow v}$ represents the final reasoning process obtained using video-text interleaving. $V_k \oplus \mathcal{R'}_t$ denotes that the key video $V_k$ is embedded within text reasoning $\mathcal{R'}_t$, enhancing the reasoning process by incorporating relevant visual context.

\section{Video-Text Interleaved Benchmark}
To fill the gap in current video understanding reasoning research, we develop a novel Video-Text Interleaved Benchmark (ViTIB). This dataset integrates video frames with corresponding text to simulate human-like video comprehension. As shown in Figure~\ref{fig:dataset}(a), ViTIB is built on VideoEspresso\cite{han2025videoespresso}, which contains 14 scenes with rich spatial details, temporal coherence, and multimodal annotations.

\subsection{Key-Video Extraction}
As illustrated in Figure~\ref{fig:dataset} (b), the process is as follows:

\subsubsection{Automatic Sample Removal}
The first step in this process is the automatic removal of incomplete or irrelevant samples. Specifically, we eliminate data points where video, question, or answer components are entirely missing. This ensures that only complete and fully coherent video-question-answer pairs are retained, which is vital for maintaining the quality and integrity of the dataset.

\subsubsection{Key-Frames Recognition}
After obtaining the samples, we extract key frames to construct representative and meaningful frame sequences. Using the powerful Gemini-2.0-Flash \cite{team2024gemini}, we identify frames most relevant to the reasoning task. Given the video, question, reasoning steps, and answer, the model selects frame indices that best support accurately answering the question.

\subsubsection{Assemlble Key-Video}
After obtaining the sequence of key-frames, we then extract the corresponding frames from the original video to assemble the key-video with a frame rate of FPS = 1. This process results in the initial Key-Video Interleaved data format.

\subsection{Human Recheck}

To ensure the quality of the benchmark and guarantee that each frame in the key-video can support reasoning to the correct answer, we conduct a thorough manual recheck, as shown in Figure~\ref{fig:dataset} (c). Specifically, each data entry is initially reviewed by three independent reviewers to verify its accuracy and relevance. To facilitate this process, we provide the reviewers with a detailed scoring guideline that outlines the evaluation criteria. The content of the scoring guideline is as follows:

\begin{datasetbox}

\begin{itemize}
    \item [\textbf{0 - 60}] The Key-Video is largely irrelevant to reasoning the correct answer and may even deviate from the topic.
    \vspace{3mm}
    \item [\textbf{60-70}] The Key-Video contains only a small number of frames that contribute to reasoning the correct answer.
    \vspace{3mm}
    \item [\textbf{70-80}] The Key-Video, in conjunction with the original video, can provide a rough for reasoning the correct answer.
    \vspace{3mm}
    \item [\textbf{80-90}] The Key-Video, when combined with the original video, allows for a complete and coherent reasoning process that leads to the correct answer.
    \vspace{3mm}
    \item [\textbf{90-100}] The correct answer can be independently reasoned from the Key-Video alone, without the need for additional context from the original video.
\end{itemize}

\end{datasetbox}

This guideline ensures that all reviewers follow a consistent and standardized approach in evaluating data, enhancing the reliability and robustness of the dataset. 
To ensure consistent quality across the three reviewers, we only retain data where all three assign scores above 80. If any entry scores below 80, the reviewers must discuss the question, reselect key-frames from the original video, and reassemble the key-video, producing a revised version. This rigorous process results in an overall average score of 83.6, ensuring the high quality of the Video-Text Interleaved Benchmark.

\begin{figure}[t]
  \centering
  \includegraphics[width=0.95\linewidth]{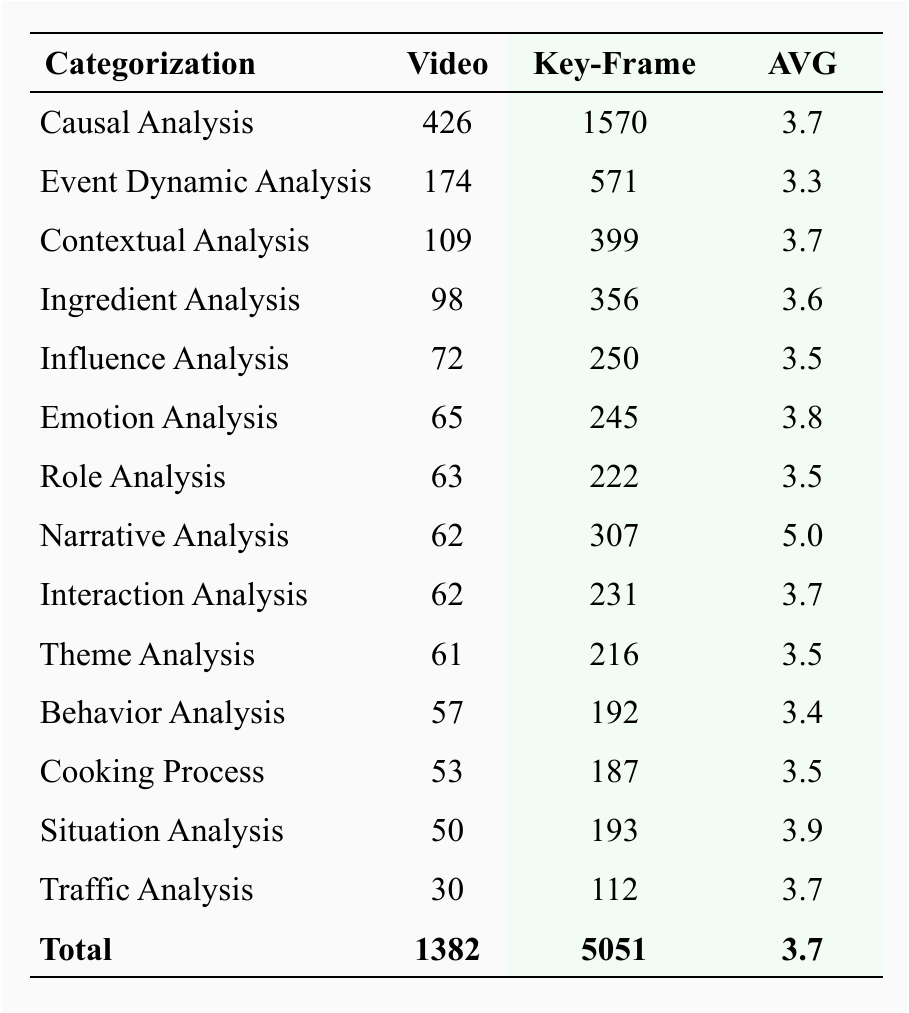}
  \caption{Display of the Number of Videos and Key Frames in the Video-Text Interleaved Benchmark. AVG represents the average number of frames per key-video.}
  \label{fig:dataset_ana}
  \vspace{-3mm}
\end{figure}

\subsection{Video-Text Interleaved Benchmark}

The Video-Text Interleaved Benchmark (ViTIB) is illustrated in Figure~\ref{fig:dataset} (d), which includes the original video, questions, options, key-videos, reasoning process, and answers. 

\subsubsection{Video-Text Interleaved Benchmark Statistics}

The detailed content of the ViTIB is shown in Figure~\ref{fig:dataset_ana}, which consists of 14 categories. These categories are carefully curated to cover a wide range of everyday scenarios, ensuring that the dataset can be applied to various reasoning tasks. Additionally, the dataset contains a total of 1,382 videos. It provides a vital and comprehensive resource for exploring and advancing Video-Text Interleaved reasoning.

In addition, the dataset includes 5,051 frames from key-videos, with an average of 3.7 frames per key-video. Notably, in the ``Narrative Analysis'', each key-video contains 5.0 frames, which aligns with the requirement for ``Narrative Analysis'' to have more key-frames as evidence to support reasoning. This demonstrates that the ViTIB effectively meets the diverse and comprehensive reasoning requirements of Video-Text Interleaved reasoning.

\subsubsection{Key-Frame Coverage of Text Reasoning}
\begin{figure}[t]
  \centering
  \includegraphics[width=\linewidth]{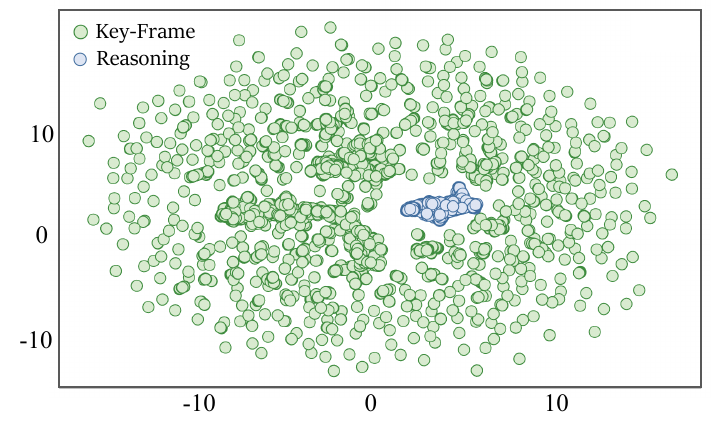}
  \vspace{-6mm}
  \caption{Utilizing CLIP \cite{radford2021learning} to visualize the semantic features of key-frames and textual reasoning content, where the semantic features of the key-frames encompass the entire scope of the reasoning content.}
  \label{fig:map}
  \vspace{-2mm}
\end{figure}

As shown in Figure~\ref{fig:map}, we employ CLIP~\cite{radford2021learning} to encode both key-frames and textual reasoning answers, then apply UMAP~\cite{mcinnes2018umap} for dimensionality reduction to visualize their features. Our analysis shows that key-frame features largely overlap with and encompass nearly the entire feature space of textual reasoning content. This indicates a strong correlation between key-frames and textual reasoning, with key-frames effectively representing the essence of the reasoning process.

\section{Video-Text Interleaved CoT Paradigm}
To address the limitations of previous video understanding methods, as shown in Figure~\ref{fig:vitcot}, we introduce a novel video understanding paradigm that applies to all CoT methods,  \titleblue{\textbf{Video}}-\titlered{\textbf{Text}} Interleaved Chain-of-Thought (\titleblue{\textbf{V}}i\titlered{\textbf{T}}CoT). Specifically, Video-Text Interleaved reasoning primarily consists of two components: Initial Text Reasoning ($\S \ref{Initial Text Reasoning}$) and Video-Text Interleaved Reasoning ($\S \ref{Video-Text Interleaved Reasoning}$).

\subsection{Initial Text Reasoning}
\label{Initial Text Reasoning}
In order to establish a foundation for Video-Text Interleaved reasoning, as shown in Figure~\ref{fig:vitcot} (b), we begin by constructing an initial framework for reasoning. Specifically, we first provide MLLMs with the \Video{[Original Video  $\mathcal{V}$]}, the \Question{[Question $\mathcal{Q}$]}, and the \Question{[Options $\mathcal{O}$]}, instructing the MLLMs to perform an \InitialReasoning{[Initial Reasoning $\mathcal{R'}$]} process without directly supplying the answer. Concretely, the specific input provided to MLLMs is as follows:

\begin{figure*}[t]
	\centering
  \includegraphics[width=\textwidth]{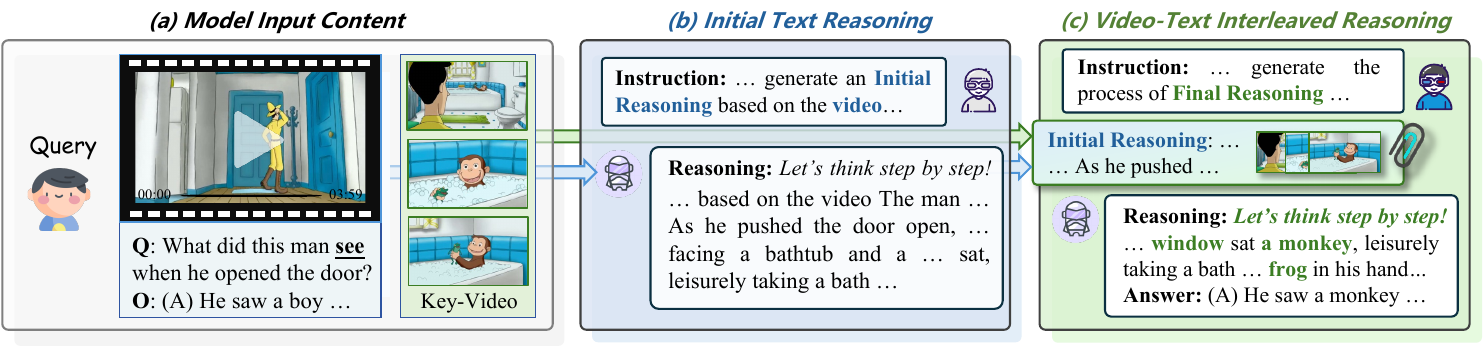}
  \vspace{-4mm}
  \caption{The overall workflow of the Video-Text Interleaved Chain-of-Thought (ViTCoT) consists of main stages: (a) \textit{Model Input Content}, (b) \textit{Initial Text Reasoning}, (c) \textit{Video-Text Interleaved Reasoning}.}
  \label{fig:vitcot}
  \vspace{-2mm}
\end{figure*}

\begin{vitcot2}
	\ \ 
	
	\textbf{Input:} \Video{[Original Video  $\mathcal{V}$]}/\Question{[Question $\mathcal{Q}$]}/\Question{[Options $\mathcal{O}$]}
	
	\textbf{Instruction:} ... You need to generate an Initial Reasoning based on the video and Options that can solve the Question, rather than directly providing the answer ...
	\\
\end{vitcot2}

\noindent
Both the \textbf{Input} and \textbf{Instruction} serve as inputs to the MLLMs, generating the \InitialReasoning{[Initial Reasoning $\mathcal{R'}$]}.
This step enables the MLLMs to analyze the video and question alongside the question and options, paving the way for more accurate reasoning.

\subsection{Video-Text Interleaved Reasoning}
\label{Video-Text Interleaved Reasoning}
Once the \InitialReasoning{[Initial Reasoning $\mathcal{R'}$]} process is obtained, as shown in Figure~\ref{fig:vitcot} (c), we start the Video-Text Interleaved Reasoning stage. Specifically, in the stage-2 the \Video{[Key-Video  $\mathcal{V}_k$]} is integrated into the \InitialReasoning{[Initial Reasoning $\mathcal{R'}$]}. Then, both the \Video{[Original Video  $\mathcal{V}$]} and the \Video{[Key-Video  $\mathcal{V}_k$]} embedded within \InitialReasoning{[Initial Reasoning $\mathcal{R'}$]} are provided together as context to the MLLMs. This approach enables the model to generate the \InitialReasoning{[Final  Reasoning $\mathcal{R}$]} process and answer by combining the video understanding with the preceding reasoning. Specifically, the input of the MLLMs is as follows:
\vspace{-1mm}

\begin{vitcot1}
	\ \ 
	
	\textbf{Input:} \Video{[Original Video  $\mathcal{V}$]}/\Question{[Question $\mathcal{Q}$]}/\Question{[Options $\mathcal{O}$]}
	
	\textbf{Instruction:} ... You need to generate the process of Final  Reasoning for the Question by integrating the \InitialReasoning{[Initial Reasoning $\mathcal{R'}$ and \Video{Key-Video  $\mathcal{V}_k$]}} process ...
	\\
\vspace{-1mm}
\end{vitcot1}
\noindent
At this stage, both the \textbf{Input} and \textbf{Instruction} serve as inputs to the MLLMs, generating the \InitialReasoning{[Final Reasoning $\mathcal{R'}$]} derived through the Video-Text Interleaved Reasoning Paradigm. 
This approach leverages Video-Text Interleaved Reasoning to integrate textual and video information for an accurate solution.

\section{Experiments}

\subsection{Vanilla CoT Reasoning}

In the main experiments, besides the \textit{Direct} queries with MLLMs, we select the following mainstream reasoning methods as baselines:
\vspace{-5mm}
\begin{vanbox}

\begin{itemize}
    \item \textit{Vanilla Chain-of-Thought} (CoT) \cite{wei2022chain}: CoT uses the prompt ``let's think step by step!'' to guide LLMs in problem-solving reasoning, thereby enhancing their reasoning capabilities.
    \vspace{1mm}
    \item \textit{Vanilla Desp-CoT} \cite{wu2023role}: Desp-CoT adopts a two-stage reasoning framework where LLMs first conduct problem analysis, then use the insights to generate the final reasoning.
    \vspace{1mm}
    \item \textit{Vanilla Plan-and-Solve} \cite{wang2023plan}: Plan-and-Solve uses a fine-grained framework that divides the problem into subproblems, which are solved individually to derive the answer.
\end{itemize}

\end{vanbox}

\subsection{Video-Text Interleaved CoT Reasoning}

To assess the effectiveness and generalizability of the Video-Text Interleaved CoT paradigm, we apply ViTCoT to current mainstream CoT methods using the Oracle key-video in the main experiments (the performance using non-Oracle key-video is presented in Section~\ref{analysis}). Specifically, we implement the following adaptations:

\begin{vitbox}

\begin{itemize}
    \item \textit{ViT Chain-of-Thought} (ViTCoT): In the first step, we prompt the LLMs to generate an initial rough reasoning. Then, the Oracle key-video is interleaved into this preliminary reasoning. Subsequently, a Video-Text Interleaving process is applied, where reasoning is conducted step by step to ultimately arrive at the final reasoning.
    \vspace{2mm}
    \item \textit{ViT Desp-CoT} \cite{wu2023role}: We directly employ the original two-stage framework of Desp-CoT. In the first stage, question analysis reasoning is generated, and the Oracle key-video is interleaved into the question analysis reasoning. In the second stage, Video-Text Interleaving is applied for problem-solving reasoning to generate the final answer.
    \vspace{2mm}
    \item \textit{ViT Plan-and-Solve} \cite{wang2023plan}: We directly utilize the original fine-grained framework of Plan-and-Solve. In the first stage, subproblems are generated, and the Oracle key-video is interleaved into these subproblems. Subsequently, fine-grained reasoning is conducted through Video-Text Interleaving to generate the final answer.
\end{itemize}

\end{vitbox}

\subsection{Experiments Setting}

We validate experiments on a series of MLLMs using the Video-Text Interleaved Benchmark. The models used include the open-source models \textit{Qwen2.5-VL-3B-Instruct}~\cite{yang2024qwen2}, \textit{Qwen2.5-VL-7B-Instruct}~\cite{yang2024qwen2}, VideoLLaMA3-7B \cite{damonlpsg2025videollama3}, \textit{Intern2.5-VL-8B} \cite{chen2024internvl}, and the closed-source model \textit{Gemini-2.0-Flash} \cite{team2024gemini}. The video input for all models is standardized to a frame rate of FPS = 1. For consistency across all experiments, the top-p and temperature parameters remain at their default values of MLLMs, typically ranging between [0, 1], ensuring that each model operates under similar conditions for a fair comparison. 
Experiments with all open-source models occur on eight NVIDIA RTX A6000 GPUs, while \textit{Gemini-2.0-Flash} uses the API calls provided by the Google AI Studio platform.

\begin{table*}[t]
\caption{The experimental results of Acc. (\%) on MLLMs.  \raisebox{0.7mm}{\colorbox{myblue}{\textcolor{myblue}{\rule{0.7mm}{0.7mm}}}} area: vanilla CoT reasoning.  \raisebox{0.7mm}{\colorbox{mygreen}{\textcolor{mygreen}{\rule{0.7mm}{0.7mm}}}} area: video-text interleaved CoT reasoning. The \underline{underline} indicates better performance in the MLLM. The \titlered{Red} represents better AVG performance in the MLLM.}
\setlength{\tabcolsep}{6pt}
\centering 
\begin{adjustbox}{width=0.99\textwidth}
\begin{tabular}{lccccccccccccccc}
\toprule 
\textbf{Methods} & \textbf{Narra.} & \textbf{Event} & \textbf{Ingre.} & \textbf{Causal} & \textbf{Theme} & \textbf{Conte.} & \textbf{Influ.} & \textbf{Role} & \textbf{Inter.} & \textbf{Behav.} & \textbf{Emoti.} & \textbf{Cook.} & \textbf{Traff.} & \textbf{Situa.} & \underline{\textcolor{titlered}{\textbf{AVG}}} \\
\midrule 
\rowcolor[rgb]{ .980, .980, .980}
\multicolumn{16}{c}{\rule{0pt}{2ex}\textit{\fontsize{11pt}{11pt} Qwen2.5-VL-3B-Instruct} \cite{yang2024qwen2}} \\
\midrule
\rowcolor[rgb]{ .970, .978, .999 }
\text{Direct} & 38.0 & 36.9 & 46.9 & 28.4 & 34.4 & 39.4 & 41.7 & 31.7 & 35.5 & 22.8 & 32.3 & 41.5 & 53.3 & 38.0 & 37.2 \\
\rowcolor[rgb]{ .970, .978, .999 }
\text{Vanilla CoT} & 34.2 & 29.3 & 37.8 & 26.1 & 36.1 & 39.4 & 36.1 & 36.5 & 30.6 & 19.3 & 32.3 & 32.1 & 46.7 & 30.0 & 33.3 \\
\rowcolor[rgb]{ .970, .978, .999 }
\text{Vanilla Desp-CoT} & 44.3 & 37.6 & 50.0 & 35.0 & 34.4 & 43.1 & 51.4 & 33.3 & 41.9 & 29.8 & 40.0 & 39.6 & 56.7 & 44.0 & 41.5 \\
\rowcolor[rgb]{ .970, .978, .999 }
\text{Vanilla Plan-and-Solve} & 49.4 & 40.8 & 58.2 & 42.3 & 47.5 & 48.6 & 48.6 & 39.7 & 50.0 & 38.6 & 41.5 & 50.9 & 63.3 & 50.0 & 47.8 \\
\midrule
\rowcolor[rgb]{ .961, .985, .970 }
\text{ViT CoT} & 43.0 & 31.2 & 40.8 & 30.5 & 37.7 & 39.4 & 41.7 & 33.3 & 33.9 & 21.1 & 32.3 & 32.1 & 43.3 & 34.0 & 35.3 \\
\rowcolor[rgb]{ .961, .985, .970 }
\text{ViT Desp-CoT} & 50.6 & 37.6 & 53.1 & 34.3 & 41.0 & 45.0 & 51.4 & 34.9 & 46.8 & 33.3 & \underline{43.1} & 47.2 & 56.7 & 46.0 & 44.4 \\
\rowcolor[rgb]{ .961, .985, .970 }
\text{ViT Plan-and-Solve} & \underline{51.9} & \underline{41.4} & \underline{62.2} & \underline{43.0} & \underline{49.2} & \underline{53.2} & \underline{52.8} & \underline{41.3} & \underline{54.8} & \underline{40.4} & \underline{43.1} & \underline{54.7} & \underline{66.7} & \underline{52.0} & \underline{\textcolor{titlered}{\textbf{50.5}}} \\
\midrule
\rowcolor[rgb]{ .980, .980, .980}
\multicolumn{16}{c}{\rule{0pt}{2ex}\textit{\fontsize{11pt}{11pt} Qwen2.5-VL-7B-Instruct} \cite{yang2024qwen2}} \\
\midrule
\rowcolor[rgb]{ .970, .978, .999 }
\text{Direct} & 49.4 & 45.2 & 58.2 & 43.4 & 44.3 & 53.2 & 47.2 & 44.4 & 38.7 & 45.6 & \underline{50.8} & 56.6 & 66.7 & 54.0 & 49.8 \\
\rowcolor[rgb]{ .970, .978, .999 }
\text{Vanilla CoT} & 53.2 & 37.6 & 45.9 & 37.3 & 44.3 & 45.9 & 41.7 & 39.7 & 41.9 & 38.6 & 43.1 & 37.7 & 46.7 & 46.0 & 42.8 \\
\rowcolor[rgb]{ .970, .978, .999 }
\text{Vanilla Desp-CoT} & 51.9 & 46.5 & 54.1 & 41.8 & \underline{47.5} & 51.4 & 47.2 & 38.1 & 40.3 & 45.6 & 46.2 & 56.6 & 70.0 & 48.0 & 48.9 \\
\rowcolor[rgb]{ .970, .978, .999 }
\text{Vanilla Plan-and-Solve} & 49.4 & 43.3 & 52.0 & 41.1 & 39.3 & 48.6 & 45.8 & 39.7 & 40.3 & 45.6 & 49.2 & 60.4 & 70.0 & 54.0 & 48.5 \\
\midrule
\rowcolor[rgb]{ .961, .985, .970 }
\text{ViT CoT} & 51.9 & 44.6 & 56.1 & 42.3 & 44.3 & \underline{56.9} & \underline{51.4} & 44.4 & \underline{46.8} & 49.1 & \underline{50.8} & \underline{66.0} & 63.3 & 52.0 & 51.4 \\
\rowcolor[rgb]{ .961, .985, .970 }
\text{ViT Desp-CoT} & 54.4 & \underline{47.8} & \underline{60.2} & 43.4 & \underline{47.5} & 56.0 & 50.0 & 41.3 & 43.5 & 52.6 & \underline{50.8} & 60.4 & \underline{76.7} & \underline{56.0} & \underline{\textcolor{titlered}{\textbf{52.9}}} \\
\rowcolor[rgb]{ .961, .985, .970 }
\text{ViT Plan-and-Solve} & \underline{57.0} & 44.6 & 58.2 & \underline{44.1} & 41.0 & 56.0 & 50.0 & \underline{47.6} & \underline{46.8} & \underline{54.4} & \underline{50.8} & 56.6 & 70.0 & 54.0 & 52.2 \\
\midrule
\rowcolor[rgb]{ .980, .980, .980}
\multicolumn{16}{c}{\rule{0pt}{2ex}\textit{\fontsize{11pt}{11pt} VideoLLaMA3-7B} \cite{damonlpsg2025videollama3}} \\
\midrule
\rowcolor[rgb]{ .970, .978, .999 }
\text{Direct} & 57.0 & 47.8 & \underline{54.1} & 49.3 & 42.6 & 62.4 & 51.4 & 46.0 & 43.5 & 47.4 & 52.3 & 62.3 & 50.0 & 58.0 & 51.7 \\
\rowcolor[rgb]{ .970, .978, .999 }
\text{Vanilla CoT} & 55.7 & 49.7 & 44.9 & 51.2 & 36.1 & 56.0 & 56.9 & 47.6 & 45.2 & 38.6 & 53.8 & 58.5 & 43.3 & 58.0 & 49.7 \\
\rowcolor[rgb]{ .970, .978, .999 }
\text{Vanilla Desp-CoT} & 49.4 & 42.7 & 44.9 & 42.7 & 49.2 & 48.6 & 50.0 & 54.0 & 45.2 & 49.1 & 50.8 & 60.4 & 53.3 & 56.0 & 49.7 \\
\rowcolor[rgb]{ .970, .978, .999 }
\text{Vanilla Plan-and-Solve} & 57.0 & 47.8 & 49.0 & 50.5 & 47.5 & 63.3 & 54.2 & 46.0 & 50.0 & 43.9 & 50.8 & 47.2 & 56.7 & 54.0 & 51.3 \\
\midrule
\rowcolor[rgb]{ .961, .985, .970 }
\text{ViT CoT} & \underline{58.2} & 45.9 & 53.1 & 51.6 & 52.5 & 58.7 & 56.9 & 46.0 & \underline{53.2} & 50.9 & 52.3 & 58.5 & 56.7 & 50.0 & 53.2 \\
\rowcolor[rgb]{ .961, .985, .970 }
\text{ViT Desp-CoT} & 50.6 & 43.9 & 46.9 & 46.5 & 55.7 & 57.8 & 45.8 & \underline{55.6} & 51.6 & 50.9 & \underline{56.9} & 64.2 & 53.3 & \underline{64.0} & 53.1 \\
\rowcolor[rgb]{ .961, .985, .970 }
\text{ViT Plan-and-Solve} & \underline{58.2} & \underline{52.9} & 53.1 & \underline{55.4} & \underline{59.0} & \underline{65.1} & \underline{58.3} & 49.2 & \underline{53.2} & \underline{56.1} & 52.3 & \underline{66.0} & \underline{60.0} & 62.0 & \underline{\textcolor{titlered}{\textbf{57.2}}} \\
\midrule
\rowcolor[rgb]{ .980, .980, .980}
\multicolumn{16}{c}{\rule{0pt}{2ex}\textit{\fontsize{11pt}{11pt} Intern2.5-VL-8B} \cite{chen2024internvl}} \\
\midrule
\rowcolor[rgb]{ .970, .978, .999 }
\text{Direct} & 67.1 & 55.4 & \underline{65.3} & 49.5 & 60.7 & 62.4 & 48.6 & 44.4 & 48.4 & 49.1 & 58.5 & 58.5 & 63.3 & 64.0 & 56.8 \\
\rowcolor[rgb]{ .970, .978, .999 }
\text{Vanilla CoT} & 64.6 & 54.1 & 58.2 & 48.4 & 55.7 & 61.5 & 47.2 & 47.6 & 41.9 & 45.6 & 49.2 & 60.4 & 56.7 & 64.0 & 53.9 \\
\rowcolor[rgb]{ .970, .978, .999 }
\text{Vanilla Desp-CoT} & 68.4 & 54.1 & 63.3 & 49.8 & 52.5 & 62.4 & 47.2 & 47.6 & 48.4 & 50.9 & 52.3 & 60.4 & 56.7 & 64.0 & 55.6 \\
\rowcolor[rgb]{ .970, .978, .999 }
\text{Vanilla Plan-and-Solve} & 59.5 & 54.8 & 62.2 & 46.9 & 55.7 & 62.4 & 48.6 & 46.0 & 41.9 & 45.6 & 49.2 & 60.4 & 60.0 & 64.0 & 54.1 \\
\midrule
\rowcolor[rgb]{ .961, .985, .970 }
\text{ViT CoT} & 59.5 & \underline{56.1} & 63.3 & 49.1 & \underline{62.3} & \underline{67.9} & \underline{52.8} & \underline{52.4} & \underline{51.6} & 49.1 & 53.8 & 58.5 & 56.7 & 56.0 & 56.4 \\
\rowcolor[rgb]{ .961, .985, .970 }
\text{ViT Desp-CoT} & \underline{69.6} & 54.1 & 62.2 & 50.0 & 57.4 & 65.1 & 50.0 & 50.8 & 46.8 & \underline{52.6} & 56.9 & 60.4 & 60.0 & 64.0 & 57.1 \\
\rowcolor[rgb]{ .961, .985, .970 }
\text{ViT Plan-and-Solve} & 68.4 & \underline{56.1} & \underline{65.3} & \underline{50.9} & 60.7 & 66.1 & 47.2 & 46.0 & 46.8 & \underline{52.6} & \underline{60.4} & \underline{62.3} & \underline{66.0} & \underline{66.0} & \underline{\textcolor{titlered}{\textbf{58.2}}} \\
\midrule
\rowcolor[rgb]{ .980, .980, .980}
\multicolumn{16}{c}{\rule{0pt}{2ex}\textit{\fontsize{11pt}{11pt} Gemini-2.0-Flash} \cite{team2024gemini}} \\
\midrule
\rowcolor[rgb]{ .970, .978, .999 }
\text{Direct} & 69.6 & 40.1 & 64.3 & 52.8 & 50.8 & 60.6 & 48.6 & 36.5 & 53.2 & 49.1 & 52.3 & 62.3 & 56.7 & 58.0 & 53.9 \\
\rowcolor[rgb]{ .970, .978, .999 }
\text{Vanilla CoT} & 60.8 & 53.5 & 57.1 & 45.8 & 50.8 & \underline{63.3} & 43.1 & 46.0 & \underline{56.5} & 52.6 & 50.8 & 52.8 & 50.0 & 52.0 & 52.5 \\
\rowcolor[rgb]{ .970, .978, .999 }
\text{Vanilla Desp-CoT} & 68.4 & 54.8 & 64.3 & 55.6 & 57.4 & 60.6 & 54.2 & 39.7 & \underline{56.5} & 43.9 & 52.3 & 60.4 & 66.7 & 64.0 & 56.9 \\
\rowcolor[rgb]{ .970, .978, .999 }
\text{Vanilla Plan-and-Solve} & 64.6 & 49.0 & 65.3 & 48.4 & 44.3 & 61.5 & 41.7 & 38.1 & 46.8 & 38.6 & 38.5 & 58.5 & \underline{70.0} & 48.0 & 51.0 \\
\midrule
\rowcolor[rgb]{ .961, .985, .970 }
\text{ViT CoT} & 64.6 & 51.6 & 60.2 & 47.7 & 50.8 & 61.5 & 40.3 & \underline{49.2} & \underline{56.5} & 47.4 & 50.8 & \underline{64.2} & 63.3 & 54.0 & 54.4 \\
\rowcolor[rgb]{ .961, .985, .970 }
\text{ViT Desp-CoT} & 69.6 & \underline{57.3} & 65.3 & \underline{56.1} & \underline{58.2} & 61.5 & \underline{55.6} & 42.9 & \underline{56.5} & \underline{57.9} & \underline{60.0} & \underline{64.2} & \underline{70.0} & \underline{66.0} & \underline{\textcolor{titlered}{\textbf{60.1}}} \\
\rowcolor[rgb]{ .961, .985, .970 }
\text{ViT Plan-and-Solve} & \underline{72.2} & 52.2 & \underline{66.3} & 50.5 & 42.6 & \underline{63.3} & 45.8 & 44.4 & 46.8 & 47.4 & 43.1 & \underline{64.2} & 63.3 & 58.0 & 54.3 \\
\bottomrule 
\end{tabular}
\end{adjustbox}
\label{main results}
\end{table*}

\subsection{Main Results}

The main experimental results are shown in Table~\ref{main results}. Based on these experimental results, we can observe that:

\textbf{1. The Video-Text Interleaved Reasoning paradigm is universally applicable across various methods.} In complex video reasoning scenarios, traditional vanilla reasoning approaches may even underperform compared to direct methods. However, methods that incorporate Video-Text Interleaved Reasoning consistently outperform base reasoning methods. On average, these methods show a significant improvement of 3.5\%. This demonstrates that the Video-Text Interleaved Reasoning paradigm significantly enhances reasoning for complex video understanding tasks. Integrating key-video with textual information results in a more intuitive and effective approach, driving performance improvements.

\

\textbf{2. The Video-Text Interleaved Reasoning paradigm is applicable to all MLLMs.} Methods incorporating Video-Text Interleaved Reasoning consistently outperform their non-Video-Text Interleaved counterparts across various models, ranging from open-source \textit{Qwen2.5-VL-3B-Instruct} to \textit{Intern2.5-VL-8B} models, as well as closed-source \textit{Gemini-2.0-Flash}. Notably, on the \textit{Qwen2.5-VL-7B-Instruct}, the video-text interleaved methods achieve an average improvement of 5.4\% over the vanilla reasoning approach. Similarly, significant gains are observed across other benchmarks, with improvements of up to 3-4\% on various tasks. These results clearly demonstrate the broad applicability and effectiveness of the Video-Text Interleaved Reasoning paradigm in enhancing the reasoning capabilities of MLLMs, even in complex video scenarios.

\begin{figure*}[t]
	\centering
  \includegraphics[width=\textwidth]{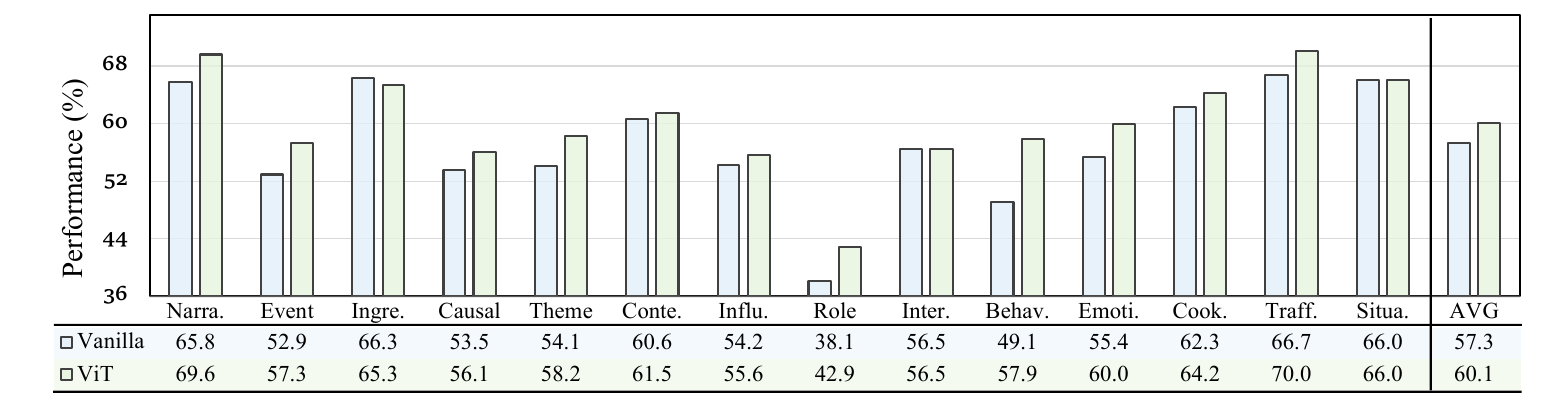}
  \caption{Compare the Video-Text Interleaved (ViT) reasoning method and Vanilla reasoning approach (with both the original video and Oracle key-video as input) on \textit{Gemini-2.0-Flash}.}
  \label{fig:two_video}
\end{figure*}

\subsection{Video-Text Interleaved Reasoning Analysis}
\label{analysis}

To eliminate the possibility that the performance gains are solely attributed to the Oracle key-video, we conduct additional experiments in this section. We compare Video-Text Interleaved reasoning with vanilla reasoning from multiple perspectives, demonstrating the advantages of the Video-Text Interleaved CoT paradigm itself.

\vspace{2mm}

\begin{figure}[t]
  \centering
  \includegraphics[width=\linewidth]{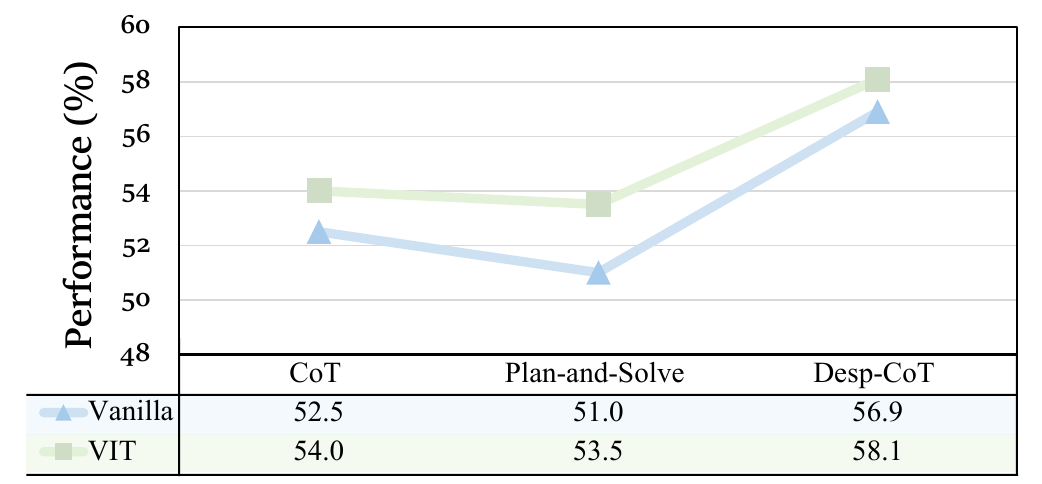}
  \caption{Comparison between the Vanilla paradigm and ViT paradigm. Video-Text Interleaved CoT Reasoning is performed using the rough key-video extracted from the original video based on the initial reasoning using CLIP~\cite{radford2021learning}.}
  \label{fig:clip_video}
\end{figure}

\textbf{1. Even when both the original and key-video are used as inputs, ViT Reasoning outperforms vanilla reasoning.}
To further validate that the Video-Text Interleaved Reasoning paradigm remains superior even when both the original video and the key video are provided as inputs, we conduct an experiment in which both videos are simultaneously fed into the vanilla reasoning methods. The results, as illustrated in Figure~\ref{fig:two_video}, show that Video-Text Interleaved Reasoning outperforms the vanilla approach by an average margin of 2.8\%. This demonstrates that, even under conditions where vanilla reasoning is enhanced with comprehensive video input, the interleaved paradigm still achieves stronger and more effective reasoning performance.

 \vspace{2mm}

\begin{figure}[t]
  \centering
  \includegraphics[width=\linewidth]{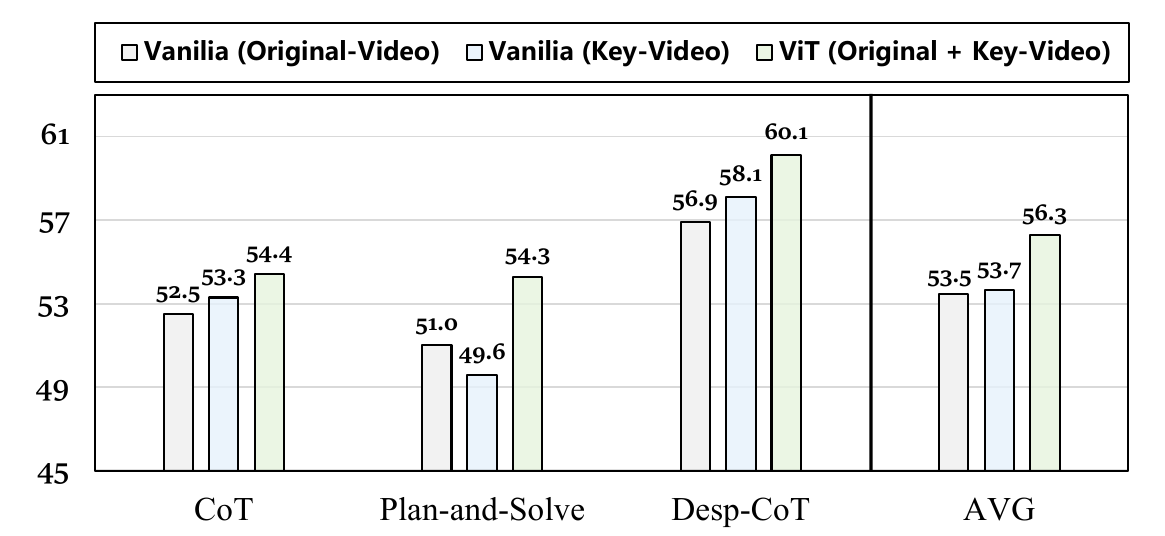}
  \caption{The performance comparison of only the Key-video input. Vanilla (Original-Video): Vanilla reasoning of only inputting Original-Video; Vanilla (Key-Video): Vanilla reasoning of only inputting Key-Video; ViT (Original+Key-Video): ViT reasoning of inputting Original-Video and Key-Video.}
  \label{fig:key_video}
  \vspace{-3mm}
\end{figure}

\begin{figure*}[t]
	\centering
  \includegraphics[width=\textwidth]{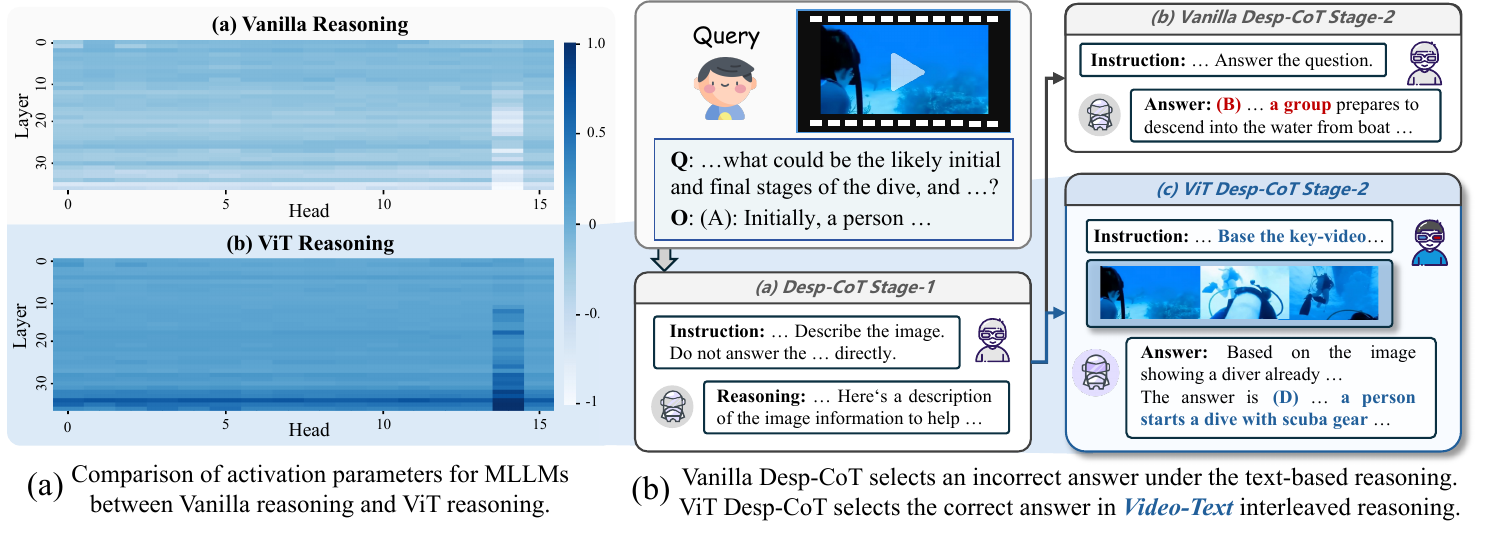}
  \vspace{-8mm}
  \caption{(a) A comparison of the number of activated neuron values between Vanilla reasoning and ViT reasoning on \textit{Qwen2.5-VL-3B}. (b) A case study of Vanilla Desp-CoT ~\cite{wu2023role} and ViT Desp-CoT on \textit{Gemini-2.0-Flash}.}
  \label{fig:case}
  \vspace{-2mm}
\end{figure*}

\textbf{2. When using non-Oracle key-video, ViT reasoning can still achieve better performance.} To verify that the performance improvement of ViT Reasoning is not solely attributed to the Oracle key-video, we validate its superiority by using a non-Oracle key-video. Specifically, after performing reasoning in stage-1 and obtaining the initial reasoning, we use CLIP~\cite{radford2021learning} to query the frames from the original video that are most similar to the initial reasoning, thereby constructing the rough key-video. We then apply Video-Text Interleaved Reasoning using the rough key-video selected by CLIP. The result is shown in Figure~\ref{fig:clip_video}, the performance of the ViT paradigm consistently surpasses that of the vanilla methods, with an average improvement of 1.7\%. This result demonstrates that ViT reasoning itself can bring performance gains to MLLMs.

 \vspace{2mm}

\textbf{3. The Video-Text Interleaved Benchmark retains rich information from the original video.} 
To evaluate the quality of the Video-Text Interleaved Benchmark, we directly compare the performance of using the key-video as the sole input with using the original video as input. The experimental results, as shown in Figure~\ref{fig:key_video}, reveal that in both the CoT and Desp-CoT, using only the key-video even outperforms using the original video, with improvements of 0.8\% and 1.2\%, respectively. This indicates that the key-video preserves the essential information of the original video while reducing irrelevant content. Furthermore, ViT reasoning remains the most optimal, confirming that it aligns with human reasoning. ViT reasonings benefit from both the original video and the key-video, preserving the detailed information in the original video while capturing the key insights from the key-video.

\textbf{4. ViT reasoning can activate more neuron values in MLLMs.} 
To better understand ViT reasoning’s effectiveness, we analyze the internal mechanisms of MLLMs by thoroughly examining activated attention heads across layers in \textit{Qwen2.5-VL-3B}. As shown in Figure~\ref{fig:case} (a), ViT reasoning triggers significantly more neuron activations than vanilla reasoning, suggesting deeper and more intensive model engagement. These results show that the ViT paradigm promotes more complex and refined reasoning in MLLMs.

 \vspace{1mm}

\textbf{5. Qualitative analysis.} To provide a clearer understanding of the workflow of ViT reasoning, we present a case study based on Desp-CoT in Figure~\ref{fig:case} (b). First, the original video, question, and options are input into the MLLM, which analyzes the video without directly answering. In (b) Vanilla Desp-CoT Stage-2, the model answers based on the first-stage analysis but selects the wrong option (\textit{B}). In contrast, in (c) ViT Desp-CoT Stage-2, the model uses video-text interleaved reasoning with the key-video and correctly selects option (\textit{D}). This case highlights the effectiveness and interpretability of ViT reasoning.

\section{Related Work}

In recent years, the rapid development of MLLMs has further advanced the field of video understanding \cite{qin2024large,wang2025multimodal, liang2024survey,zhao2023survey,fei2024enhancing}. An increasing number of researchers are applying MLLMs to video understanding tasks \cite{feng2025video,zhao2023learning,yang2023vid2seq,fei2024dysen,ji2023partial}. Specifically, Fei et al. \cite{fei2024video} introduce the concept of Video-of-Thought, which extends the CoT method by decomposing complex tasks into simpler subproblems and addressing them progressively, from pixel-level perception to high-level reasoning. Wang et al. \cite{wang2024videocot} present the VideoCoT benchmark, aiming to explore the potential limits of the reasoning abilities MLLMs that rely on textual data. Yue et al. \cite{yue2025find} propose a method for storing and retrieving video texts as dense vectors, allowing for the extraction of detailed information from lengthy videos, thereby enhancing textual reasoning and improving effectiveness. Han et al. \cite{han2025videoespresso} propose VideoEspresso, featuring semantically-aware VideoQA pairs and a Hybrid LVLMs Collaboration framework for adaptive frame selection and CoT-based video reasoning. Hu et al. \cite{hu2025cos} propose Chain-of-Shot, optimizing shot selection for long videos by aligning task-relevant and irrelevant shots, enhancing video understanding, and mitigating context length issues. Himakunthala et al.~\cite{himakunthala2023let} propose a keyframe-based video reasoning approach with two novel tasks to evaluate reasoning capabilities.

Compared to the traditional text-based reasoning paradigms, we propose for the first time the Video-Text Interleaved reasoning paradigm, which enables more intuitive reasoning by integrating video and text. Additionally, to address the limitations of previous works, we construct the Video-Text Interleaved Benchmark. This dataset serves as a foundation for testing and evaluating the efficacy of the Video-Text Interleaved reasoning paradigm.

\section{Conclusion}
In this work, we introduce a novel reasoning paradigm for video understanding, Video-Text Interleaved CoT (ViTCoT), which offers a more natural and intuitive approach aligned with human cognition by integrating key-video into the reasoning. Specifically, we construct the first Video-Text Interleaved Benchmark (ViTIB) and conduct an extensive exploration of the potential of ViTCoT. Extensive experiments demonstrate that ViT Reasoning substantially outperforms traditional text-only CoT approaches in complex video understanding tasks, activating more neuronal values. Integrating visual information with textual reasoning facilitates a more intuitive method for video understanding and reasoning, enabling MLLMs to better comprehend the complexity of the real world.

\section*{Acknowledgments}
This work was supported by the National Natural Science Foundation of China (NSFC) via grant 62306342, 62236004, 62206078 and 62476073. This work was supported by the Scientific Research Fund of Hunan Provincial Education Department (24B0001). This work was sponsored by the Excellent Young Scientists Fund in Hunan Province (2024JJ4070), the Science and Technology Innovation Program of Hunan Province under Grant 2024RC3024 and CCF-Zhipu Large Model Innovation Fund (NO.CCF-Zhipu202406).
This work was carried out in part using computing resources at the High Performance Computing Center of Central South University.

\bibliographystyle{ACM-Reference-Format}
\bibliography{sample-base}

\end{document}